\pdfoutput=1

\documentclass[11pt]{article}

\usepackage{acl}

\usepackage{times}
\usepackage{latexsym}

\usepackage[T1]{fontenc}

\usepackage[utf8]{inputenc}

\usepackage{microtype}

\usepackage{arydshln}
\usepackage{amsmath}
\usepackage{bbm}
\usepackage{bm}
\usepackage{booktabs}
\usepackage{subfigure}
\usepackage{multirow}
\usepackage{color}
\usepackage{graphicx}
\usepackage{makecell}
\usepackage{amssymb}
\usepackage{tikz}
\usepackage{pgfplots}

\title{Retaining Key Information under High Compression Ratios: Query-Guided Compressor for LLMs}

\author{Zhiwei Cao\textsuperscript{1,3}\footnotemark[1],~~\textbf{Qian Cao}\textsuperscript{2}\footnotemark[1],~~\textbf{Yu Lu}\textsuperscript{2},~~\textbf{Ningxin Peng}\textsuperscript{2},~~\textbf{Luyang Huang}\textsuperscript{2} \\
\textbf{Shanbo Cheng}\textsuperscript{2}\footnotemark[2]~~and~~\textbf{Jinsong Su}\textsuperscript{1,3}\footnotemark[2] \\
\textsuperscript{1}School of Informatics, Xiamen University~~~\textsuperscript{2}ByteDance Research \\
\textsuperscript{3}Shanghai Artificial Intelligence Laboratory \\
\small{lines1@stu.xmu.edu.cn~~\{caoqian.95, luyu.ly, chengshanbo\}@bytedance.com~~jssu@xmu.edu.cn}
}
\begin{document}
\maketitle

\renewcommand{\thefootnote}{\fnsymbol{footnote}}
\footnotetext[1]{These authors contributed equally. This work was done when Zhiwei Cao was interning at ByteDance.}
\footnotetext[2]{Corresponding author.}
\renewcommand{\thefootnote}{\arabic{footnote}}

\begin{abstract}
The growing popularity of Large Language Models has sparked interest in context compression for Large Language Models (LLMs). However, the performance of previous methods degrades dramatically as compression ratios increase, sometimes even falling to the closed-book level. This decline can be attributed to the loss of key information during the compression process. Our preliminary study supports this hypothesis, emphasizing the significance of retaining key information to maintain model performance under high compression ratios. As a result, we introduce Query-Guided Compressor (QGC), which leverages queries to guide the context compression process, effectively preserving key information within the compressed context. Additionally, we employ a dynamic compression strategy. We validate the effectiveness of our proposed QGC on the Question Answering task, including NaturalQuestions, TriviaQA, and HotpotQA datasets. Experimental results show that QGC can consistently perform well even at high compression ratios, which also offers significant benefits in terms of inference cost and throughput\footnote{Our code is available at \url{https://github.com/DeepLearnXMU/QGC}.}.

\end{abstract}

\section{Introduction}
The emergence of chatGPT~\cite{ouyang2022training} and GPT4~\cite{OpenAI_2023}, along with other Large Language Models (LLMs)~\cite{touvron2023llama,touvron2023llama2} has sparked a global sensation. The success of LLMs is closely tied to the long context capabilities of LLMs~\cite{dong2022survey,lewis2020retrieval}, especially in the field of multi-document question answering.
However, the utilization of long context also introduces challenges such as higher inference cost, longer latency, and inferior performance caused by redundant information~\cite{jiang2023longllmlingua}.

Many efforts have been made to compress the long context by directly removing a certain percentage of less important words, such as LongLLMLingua~\cite{jiang2023longllmlingua} and Selective-Context~\cite{li2023compressing}. 
Another common method is to generate a text summary of the given context~\cite{xu2023recomp,wang2023learning}.
Unlike deleting or reordering the word in the context, AutoCompressor~\cite{chevalier2023adapting} compresses long documents into multiple vectors as soft prompts, which are optimized with full parameters of LLMs.
However, our preliminary study shows that these methods have a common flaw: as the compression ratio increases, the compressed context fails to retain key information, resulting in a significant decrease in the performance of LLMs.

The key to solve this problem is query, which defines what key information is.
We aim to preserve this query-related key information even at a high compression ratio. 
Specifically, we propose the Query-Guided Compressor (QGC) to fully utilize query information throughout each compression step.
We first feed the query and the documents together into a context encoder to learn the query-guide document representations. We then compress these document representations into $n$-gram representations guided by the importance of each word in relation to the query. 
Subsequently, we propose to augment the $n$-gram representations by reviewing the query and document, which are finally aligned to the embedding space of the LLMs. 
We further propose dynamically adjusting the compression ratio of each document based on its relevance to the query.
Compared to previous methods, QGC has several advantages: 1) high compression ratios by retaining most query-related information during compression, 2) low training costs by optimizing the compressor only instead of finetuning the entire LLM, and 3) better semantic consistency by compressing the $n$-gram structure rather than deleting words.

We validate the effectiveness of QGC on the multi-document Question Answering task, including three datasets: NaturalQuestions, TriviaQA, and HotpotQA. 
Experimental results on the QA task indicate that, compared to LongLLMLingua, QGC exhibits a 2.75 times higher compression ratio and a 2.42 times higher throughput. Additionally, its accuracy has improved by an average of 5 points.
We further investigated the loss of key information throughout the compression process. The findings reveal that under high compression ratios and high noise conditions, QGC only incurs a performance loss of about 10\%, while LongLLMLingua suffers a loss of approximately 47\%. This validates the effectiveness of QGC in retaining key information.

\section{Preliminary Study}
In this section, we first briefly formulate the long context compression on the Question Answering task, and then present an analysis on the key information loss in previous compression methods.

\subsection{Task Formulation}
Given a LLM input with augmented context $\mathbf{x} = (\mathbf{x}^{ins}, \mathbf{x}^{d_{1}}, ..., \mathbf{x}^{d_{k}}, ..., \mathbf{x}^{d_{K}}, \mathbf{x}^{q})$, which consists of the instruction $\mathbf{x}^{ins}$, $K$ documents $\{\mathbf{x}^{d_{k}}\}^{K}_{k=1}$, and the query $\mathbf{x}^{q}$, the objective of context compression can be formulated as:
\begin{align}
\label{eq:task-objective}
\min_{\mathbf{\widetilde{x}}} d(\text{LLM}(\mathbf{y}|\mathbf{x}), \text{LLM}(\mathbf{\widetilde{y}}|\mathbf{\widetilde{x}})),
\end{align}

\noindent where $\mathbf{y}$ is the ground-truth answer and $\mathbf{\widetilde{y}}$ represents the output of the LLM with the compressed context $\mathbf{\widetilde{x}}$ as the input. $d(\cdot,\cdot)$ is a function measuring the distance between two distributions, such as KL divergence. In this work, we focus on compressing $K$ retrieved documents that greatly determine the length of the input.

\begin{figure}[t!]
\centering
\subfigcapskip=-2.5mm
\subfigure[Compression Ratio for LongLLMLingua]{
    \includegraphics[width=0.92\linewidth]{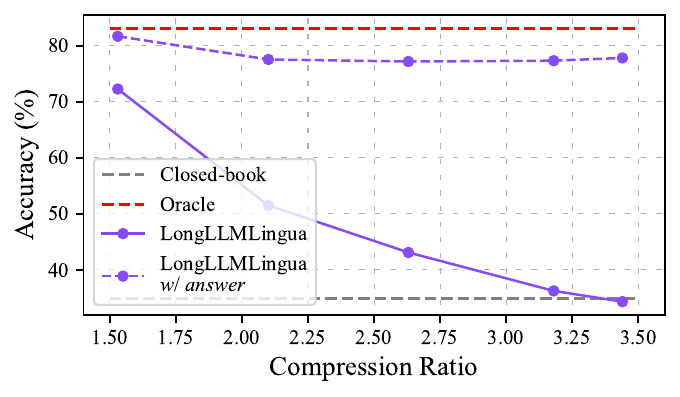}
    \label{fig:preliminary-study-longllmlingua}
}

\vspace{-2mm}
\subfigure[Document Number for AutoCompressor]{
    \includegraphics[width=0.92\linewidth]{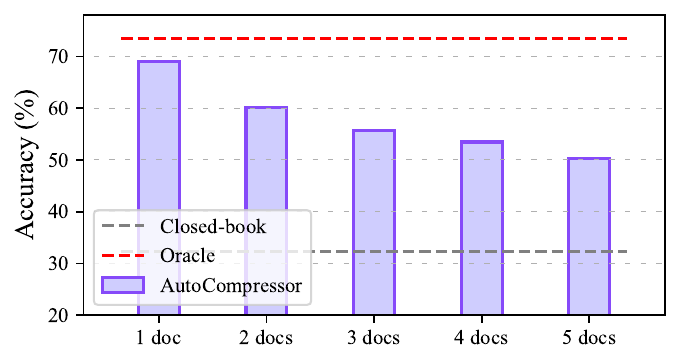}
    \label{fig:preliminary-study-autocompressor}
}
\vspace{-2mm}
\caption{The accuracy of LongLLMLingua~\cite{jiang2023longllmlingua} and AutoCompressor~\cite{chevalier2023adapting} with different compression ratios and number of documents on the NaturalQuestions test set, respectively. Closed-book denotes providing LLMs with the question only, and Oracle means using the question and corresponding ground-truth documents as the input of the LLM. ``\textit{w/ answer}'' means adding the golden answer to the compressed context.}
\label{fig:preliminary-study}
\end{figure}

\subsection{Key Information Loss in Compression}
\label{sec:preliminary-study-key-info}
We study the effectiveness of two representative methods, LongLLMLingua~\cite{jiang2023longllmlingua} and AutoCompressor~\cite{chevalier2023adapting}. We conduct experiments on the NaturalQuestions dataset~\cite{liu2023lost} and use accuracy as the evaluation metric, which judges whether any correct answers appear in the LLM prediction.

For LongLLMLingua, we apply LLaMA-2-7B-Chat\footnote{https://ai.meta.com/llama/} as the small language model for compression, and use LongChat-13B-16K\footnote{https://huggingface.co/lmsys/longchat-13b-16k} as the target LLM. We use the open-source AutoCompressor\footnote{https://github.com/princeton-nlp/AutoCompressors}, which fine-tunes LLaMA-2-7B to compress context and generate answers. Here, we consider four settings: 
\begin{itemize}
    \item \textbf{Closed-book}. It takes the query as the LLM input with no additional documents.
    \item \textbf{Oracle}. The query and only the document containing the ground truth are used as inputs to the LLM.
    \item \textbf{Base}. Based on Oracle, we compress the document directly with various compression ratios for LongLLMLingua. However, since AutoCompressor is set to compress documents to fixed length vectors, we change the compression ratio by adding external documents.
    \item \textbf{Base \textit{w/ answer}}. We manually add key information to the compressed results by concatenating the answer with the compressed word sequence in LongLLMLingua. Note that this setting is impractical for AutoCompressor where the compressed results are vectors that cannot be changed directly. 
\end{itemize}

From Figure~\ref{fig:preliminary-study}, we find that the performance of both methods degrades significantly with increasing compression ratios. As shown in Figure~\ref{fig:preliminary-study-longllmlingua}, the performance of LongLLMLingua decreases by 47\% as the compression ratio increases from 1.53x to 3.44x. Even worse, the accuracy of LongLLMLingua at 3.44x compression ratio is equivalent to that of the closed-book setting. The same findings are illustrated in Figure~\ref{fig:preliminary-study-autocompressor} for AutoCompressor. 

More importantly, we observe that adding key information to the compressed result can greatly alleviate the performance degradation that typically occurs at high compression ratios. Back to Figure~\ref{fig:preliminary-study-longllmlingua}, the accuracy line fluctuates little as the compression ratio increases from 1.5x to 3.5x with the help of additional key information, which is a decrease of 3.87\% compared to the former 47\% with the loss of key information. These observations validate the need to preserve key information during compression, which motivates us to explore a better method to fully exploit query information for context compression.

\section{Query-Guided Compression}
\begin{figure}[t!]
\centering
\includegraphics[width=0.85\linewidth]{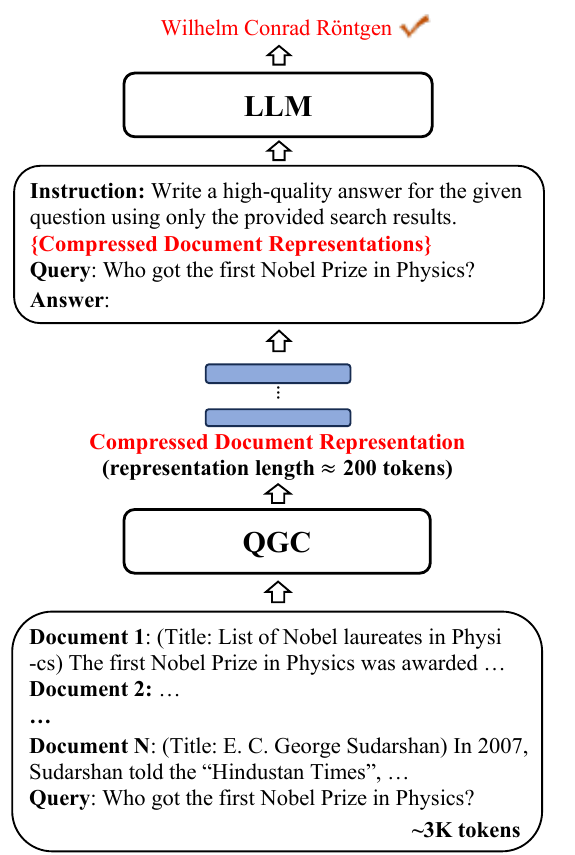}
\caption{The framework of our method.}
\label{fig:framework}
\end{figure}

As shown in Figure~\ref{fig:framework}, we equip the LLM with the Query-Guided Compressor to compress long documents into a much shorter sequence of continuous representations, which are then concatenated with the corresponding instruction and query as the input for the LLM. In the following, we first introduce the architecture of Query-Guided Compressor and then its training objective. Then, we propose a dynamic compression strategy that assigns higher compression ratios for irrelevant documents to further improve the compressed representations.

\subsection{Compressor Architecture}
\begin{figure}[t!]
\centering
\includegraphics[width=0.98\linewidth]{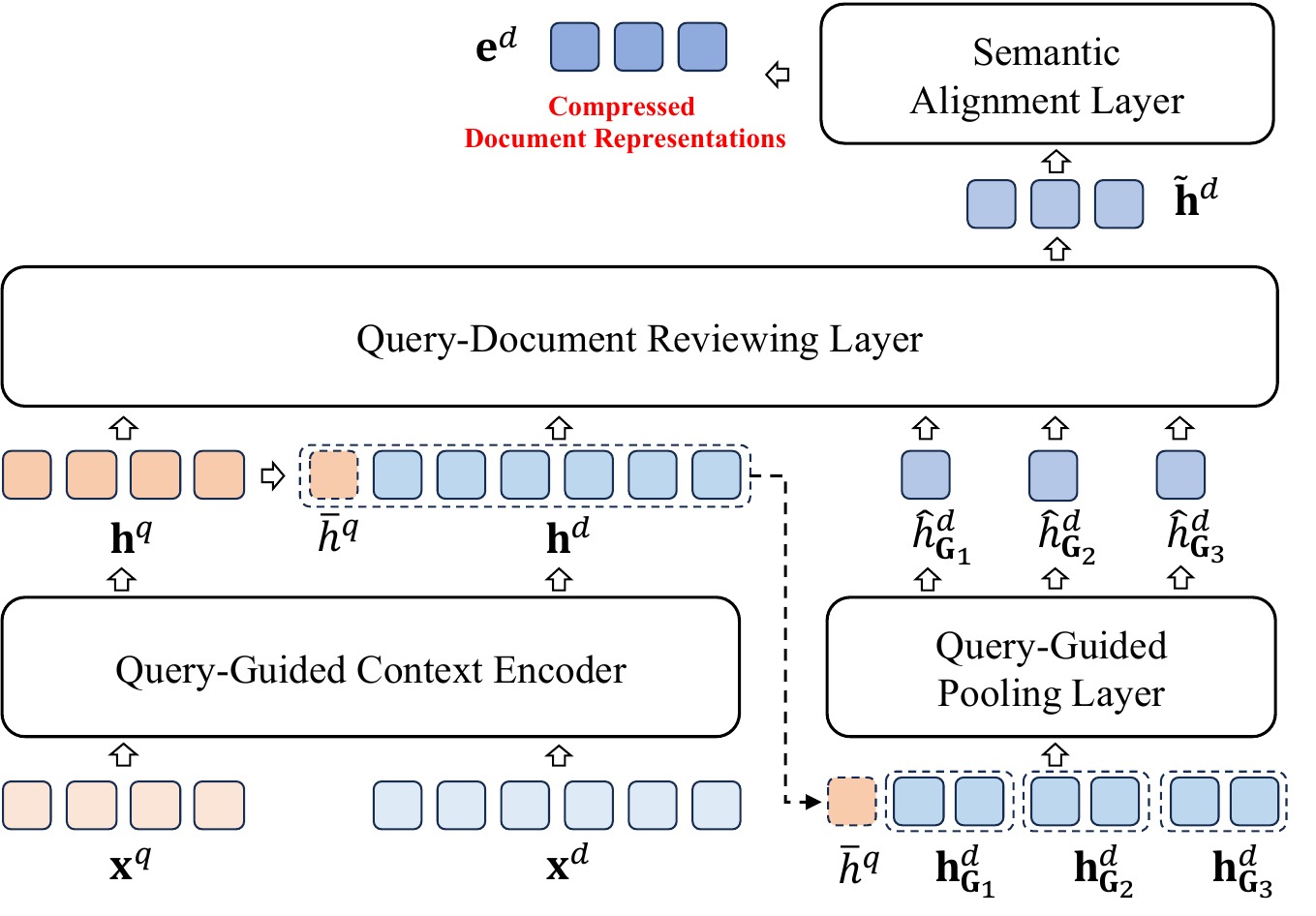}
\vspace{-1mm}
\caption{The structure of QGC. The first three layers use query $q$ to guide document $d$ encoding, pooling, and reviewing respectively. The last layer aligns document representations into the target LLM embedding space.}
\label{fig:compressor}
\end{figure}

Figure \ref{fig:compressor} illustrates the basic architecture of our Query-Guided Compressor. Using the compressor, we adopt the following steps to produce compressed representations of each document: 1) learning the query-aware document representations; 2) compressing the document representations into $n$-gram representations by weighted pooling; 3) augmenting the $n$-gram representations by reviewing the query and the entire document; 4) aligning the obtained representations into the embedding space of the LLM. Particularly, these four steps correspond exactly to the four key components of our compressor, which are all boxed in Figure~\ref{fig:compressor}. Note that we perform the above operations on each document, thus omitting the index $k$ of the document for simplicity.

\paragraph{Query-Guided Context Encoder}
At the first step, we feed the concatenation of the query $\mathbf{x}^{q}$ and the document $\mathbf{x}^{d}$ into query-aware context encoder to learn the representations of the query and the document. 

The encoder consists of two Transformer encoder layers. Formally, these representations can be obtained in the following way:
\begin{align}
    [\mathbf{h}^{q}; \mathbf{h}^{d}] = \text{ContextEncoder}([\mathbf{x}^{q};\mathbf{x}^{d}]).
\end{align}
Here, $\mathbf{h}^{q}$$=$$\{h^{q}_i\}^{N_q}_{i=1}$ and $\mathbf{h}^{d}$$=$$\{h^{d}_i\}^{N_d}_{i=1}$ are the corresponding representation sequences of the query and the document with the lengths of $N_q$ and $N_d$, respectively.
By allowing the query and the document to see each other during encoding, we can facilitate the extraction of the key information relevant to the query in the document.

\paragraph{Query-Guided Pooling Layer}
In the next step, we split the entire document into several $n$-grams and compress the information of each $n$-gram into a vector based on their correlation to the query. 

To this end, document representations are organized as follows:
\begin{align}
    \mathbf{h}^{d} &= [\mathbf{h}^{d}_{\mathbf{G}_1}, ..., \mathbf{h}^{d}_{\mathbf{G}_j}, ..., \mathbf{h}^{d}_{\mathbf{G}_{N_g}}] \\
    &= [\mathbf{h}^{d}_{1:n}, ..., \mathbf{h}^{d}_{(j-1)\times n: j\times n}, ..., \mathbf{h}^{d}_{N_d-n+1:N_d}],
    \nonumber
\end{align}
\noindent where $\mathbf{G}_j$ represent the indices of the $j$-th $n$-gram. $N_g$$=$$\frac{N_d}{n}$ is the number of $n$-grams.

Then, we measure the weight of each token in $\mathbf{G}_j$ by calculating its relevance with the mean representation $\overline{h}^{q}$ of query tokens:
\begin{align}
    \overline{h}^{q}&=\frac{1}{N_q}\sum{h^{q}_i}, \\
    w_{i,\mathbf{G}_j} &= \frac{\exp{s(\overline{h}^{q}, h^{d}_i)}}{\sum_{i^{'} \in \mathbf{G}_j} \exp{s(\overline{h}^{q}, h^{d}_{i^{'}})}},
\end{align}

\noindent where $s(\cdot, \cdot)$ is the dot-product function, and $w_{i,G_j}$ represents the weight of the $i$-th token representation $h^{d}_i$ in the document, which belongs to the $j$-th $n$-gram.

Finally, we acquire the compressed $n$-gram representations $\hat{h}^{d}_{\mathbf{G}_j}$ as the weighted sum of token representations in the $n$-gram:
\begin{align}
    \hat{h}^{d}_{\mathbf{G}_j} = \sum_{i \in \mathbf{G}_j} w_{i,\mathbf{G}_j} \cdot h^{d}_i.
\end{align}

\paragraph{Query-Document Reviewing Layer} To further prevent the key information loss in compression, we introduce a novel reviewing module to perfect the compressed $n$-gram representations by revising both the query and the document representations.

Concretely, this encoder consists of two Transformer encoder layers, which takes the query representations $\mathbf{h}^{q}$, the document representations $\mathbf{h}^{d}$, and the compressed $n$-gram representations $\hat{\mathbf{h}}^{d}$ as inputs, and outputs the improved document $n$-gram representations $\widetilde{\mathbf{h}}^{d}$:
\begin{align}
    \widetilde{\mathbf{h}}^{d} = \text{ReviewingLayer}([\mathbf{h}^{q}; \mathbf{h}^{d}; \hat{\mathbf{h}}^{d}]).
\end{align}

\paragraph{Semantic Alignment Layer}
Since $\widetilde{\mathbf{h}}^{d}$ lie in a different embedding space with the inputs of the LLM, we use a fully-connected semantic alignment layer to map the $n$-gram representations into the embedding space of the LLM. The aligned $n$-gram representations $\mathbf{e}^{d}$ can be formulated as follows:
\begin{align}
    \mathbf{e}^{d} = \textbf{W} \cdot \widetilde{\mathbf{h}}^{d} + \textbf{b},
\end{align}
\noindent where $\textbf{W}$ and $\textbf{b}$ are learnable parameters.

\subsection{Compressor Training}
\label{sec:training}
Unlike AutoCompressor~\cite{chevalier2023adapting}, we fix the parameter of the LLM and only fine-tune the compressor.

Through the above steps, each long document is compressed into a shorter sequence of continuous representations $\mathbf{e}^d$. Thus, the inputs of the LLM are finally formated as $\widetilde{\mathbf{x}} = (\mathbf{x}^{ins}, \mathbf{e}^{d_{1}}, ..., \mathbf{e}^{d_{k}}, ..., \mathbf{e}^{d_{K}}, \mathbf{x}^{q})$. To avoid missing the key information during compression, we define the training objective of the compressor in the following way:
\begin{align}
\label{eq:loss}
\mathcal{L} &= \mathcal{L}_{CE} + \mathcal{L}_{KL} \\
&= -\log{p(\mathbf{y}|\widetilde{\mathbf{x}})} + \text{KL}[p(\mathbf{y}|\mathbf{x})||p(\mathbf{y}|\widetilde{\mathbf{x}})],
\nonumber
\end{align}

\noindent where $\text{KL}[\cdot||\cdot]$ represents the Kullback–Leibler divergence. By introducing the KL loss, we encourage the LLM to generate the correct answer even with compressed representations as input.

\subsection{Dynamically Compressing Strategy}
\label{sec:inference}
Due to the different importance of retrieved documents, we propose to dynamically adjust the compression ratios for different retrieved documents. Specifically, we assign the $n$-gram size $n_k$ for the $k$-th document based on the importance ranking:
\begin{align}
\label{eq:compression-rate}
    n_{k} = 
    \begin{cases}
        \min (2 \cdot O_k, 16) & S_{k} \geq \epsilon\\
        \infty & S_{k} < \epsilon
    \end{cases},
\end{align}
\noindent where $S_k$ and $O_k$ is the score and rank of the $k$-th document acquired by the existing reranker, such as Contriever~\cite{izacard2022unsupervised}. $\epsilon$ is the score threshold for filtering low-score documents. Note that when the assigned $n$-gram size $n_{k}$ is set to $\infty$, the corresponding document will be discarded.

\begin{table*}[ht!]
\fontsize{10}{12}\selectfont
\setlength{\tabcolsep}{1.1mm}
\centering
\begin{tabular}{lccccccccc}
\toprule
\multirow{2}{*}{\textbf{Methods}} & \multicolumn{3}{c}{\textbf{NaturalQuestions}} & \multicolumn{3}{c}{\textbf{TriviaQA}} & \multicolumn{3}{c}{\textbf{HotpotQA}} \\
\cmidrule(r){2-4} \cmidrule(r){5-7} \cmidrule(r){8-10}
& Acc & CR & TP & EM & CR & TP & F1 & CR & TP \\
\midrule
\midrule
\multicolumn{10}{l}{{\textbf{LongChat-13B}}} \\
\midrule
Closed-book & 34.84 & - & - & 36.07 & - & - & 22.19 & - & - \\
Oracle & 83.05 & 59.2x & - & - & - & - & 60.61 & 42.2x & - \\
Original Prompt & 53.11 & 1.0x & - & 48.70 & 1.0x & - & 44.76 & 1.0x & - \\
\midrule
\multicolumn{10}{l}{\textit{Reranker-based Methods}} \\
Sentence-BERT~\cite{reimers-2020-multilingual-sentence-bert} & 60.75 & 4.1x & 0.137 & 48.89 & 4.5x & \textbf{1.957} & 42.92 & 4.4x & \textbf{1.930} \\
BGE-Reranker~\cite{bge_embedding} & 64.33 & 4.1x & 0.138 & 47.71 & 4.5x & 1.724 & 47.96 & 4.4x & 1.689 \\
Cond.PPL~\cite{jiang2023longllmlingua} & 65.91 & 4.1x & 0.128 & 52.48 & 4.5x & 1.287 & 49.82 & 4.3x & 1.267 \\
\midrule
\multicolumn{10}{l}{\textit{Compression-based Methods}} \\
Selective-Context~\cite{li2023compressing} & 35.44 & 2.5x & 0.077 & 42.73 & 2.5x & 0.465 & 29.68 & 2.6x & 0.456 \\
LongLLMLingua~\cite{jiang2023longllmlingua}$\dagger$ & 66.70 & 3.9x & - & - & - & - & - & - & -  \\
LongLLMLingua~\cite{jiang2023longllmlingua} & 67.01 & 4.1x & 0.118 & 51.51& 3.7x & 0.724 & 45.43 & 3.8x & 0.683 \\
\midrule
QGC & \textbf{69.19} & \textbf{15.2x} & \textbf{0.356} & \textbf{57.72} & \textbf{7.9x} & 1.832 & \textbf{52.12} & \textbf{8.8x} & 1.849 \\
\midrule
\midrule
\multicolumn{10}{l}{{\textbf{LLaMA-2-7B}}} \\
\midrule
Closed-book & 32.35 & - & - & 30.70 & - & - & 10.54 & - & - \\
Oracle & 73.45 & 59.2x & - & - & - & - & 57.68 & 42.2x & - \\
Original Prompt & 27.53 & 1.0x & - & 49.47 & 1.0x & - & 44.24 & 1.0x & - \\
\midrule
\multicolumn{10}{l}{\textit{Reranker-based Methods}} \\
Sentence-BERT~\cite{reimers-2020-multilingual-sentence-bert} & 24.26 & 4.1x & 0.133 & 49.49 & 4.5x & 0.731 & 40.65 & 4.4x & 0.752 \\
BGE-Reranker~\cite{bge_embedding} & 25.08 & 4.1x & 0.130 & 48.69 & 4.5x & 0.683 & 46.13 & 4.4x & 0.724 \\
Cond.PPL~\cite{jiang2023longllmlingua} & 27.87 & 4.1x & 0.123 & 52.76 & 4.5x & 0.602 & 47.84 & 4.3x & 0.623 \\
\midrule
\multicolumn{10}{l}{\textit{Compression-based Methods}} \\
Selective-Context~\cite{li2023compressing} & 31.79 & 2.6x & 0.082 & 48.55 & 2.5x & 0.303 & 28.21 & 2.6x & 0.332 \\
LongLLMLingua~\cite{jiang2023longllmlingua} & 41.13 & 4.1x & 0.108 & 50.44 & 3.7x & 0.432 & 39.87 & 3.8x & 0.438 \\
AutoCompressor~\cite{chevalier2023adapting} & 49.23 & 13.9x & 0.302 & 29.17 & 8.7x & 0.823 & 29.02 & 8.1x & 0.833 \\
ICAE~\cite{ge2023context} & 53.34 & \textbf{21.5x} & - & 48.91 & 10.2x & - & 34.50 & 9.5x & - \\
\midrule
QGC & \textbf{60.90} & 15.2x & \textbf{0.313} & \textbf{57.46} & 7.9x & \textbf{0.902} & \textbf{51.64} & 8.8x & \textbf{0.927} \\
QGC($\epsilon=0.42$) & 57.62 & 20.6x & - & 57.11 & \textbf{10.9x} & - & 51.23 & \textbf{12.1x} & - \\
\bottomrule
\end{tabular}
\vspace{-1mm}
\caption{Experimental results on three benchmark datasets. \textbf{Acc} = accuracy, \textbf{EM} = exact match, \textbf{F1} = F1 score, \textbf{CR} = compression ratio, \textbf{TP} = throughput (examples/second). \textbf{Closed-book}, \textbf{Oracle}, and \textbf{Original Prompt} denote using the query only, the complete ground-truth documents, and all retrieved documents as inputs, respectively. $\dagger$ indicates that the results are directly cited from \citet{jiang2023longllmlingua}.}
\vspace{-3mm}
\label{tab:main-exp}
\end{table*}

\section{Experiments}
In this section, we conduct extensive experiments to investigate the effectiveness of QGC.

\paragraph{Datasets \& Evaluation Metric}
The experiments are carried out based on the three datasets:
\begin{itemize}
    \item \textbf{NaturalQuestions} We select the processed version~\cite{liu2023lost} where each question has 20 related documents and only one of them contains the correct answer. We follow \citet{liu2023lost} to use accuracy (Acc) as the evaluation metric, which judges whether the correct answer appears in the prediction.
    \item \textbf{TriviaQA} We employ the adversarial Contriever~\cite{izacard2022unsupervised} to retrieve the top 10 documents from all Wikipedia passages.
    Following~\citet{lewis2020retrieval}, we use the Exact Match (EM) metric to evaluate the LLM prediction.
    \item \textbf{HotpotQA} Different from the above two datasets, HotpotQA~\cite{yang-etal-2018-hotpotqa} is a multi-hop dataset where the answer lies in more than one document. Specifically, each question has 10 related documents and two of them are ground-truth documents. Following \citet{yang-etal-2018-hotpotqa}, we use the F1 score to measure the correctness of the LLM.
\end{itemize}

Besides, we calculate the compression ratio (CR) for different methods, which is defined as the length rate of the original context to the compressed context. We also provide the inference throughput (TP) on a single A100-80G GPU, including compression and generation. 

\paragraph{Baselines}
Following~\cite{jiang2023longllmlingua}, we include two sets of methods as our baselines. 

1) \textit{Reranker-based Methods}. It simply uses a reranker method to sort documents based on importance and discards unimportant ones. We select the following reranker: Sentence-BERT~\cite{reimers-2020-multilingual-sentence-bert}, BGE-Reranker~\cite{bge_embedding}, and Cond.PPL proposed by~\citet{jiang2023longllmlingua} to measure the association between the query and documents. Then, we discard documents with low association until the compression ratio is met and sort the remaining documents according to the association from high to low. 

2) \textit{Compression-based Methods}. Compared with reranker-based methods, they further compress the sorted documents, retaining more information while satisfying a higher compression ratio. We select the following methods as our baselines:
\begin{itemize}
    \item \textbf{Selective-Context}~\cite{li2023compressing} It uses self-information estimated by an external language model to prune redundant words.
    \item \textbf{LongLLMLingua}~\cite{jiang2023longllmlingua} It is the state-of-the-art method for long context compression. It first uses a language model to quantify the importance of each document as its question-aware perplexity, and then designs a question-aware coarse-to-fine compression method to delete unimportant tokens.
    \item \textbf{AutoCompressor}~\cite{chevalier2023adapting} It fine-tunes LLaMA-2-7B to recursively compress long context into summary vectors, which are used as soft prompts to generate the answer. We use the released AutoCompressor-Llama-2-7B-6K for experiments.
    \item \textbf{ICAE}~\cite{ge2023context} Similar to AutoCompressor, it generates compact and informative memory slots to represent the original context. We use the released ICAE model pre-trained on Llama-2-7B-Chat for experiments~\footnote{https://github.com/getao/icae}.
\end{itemize}

\paragraph{Implementation Details}
We use LongChat-13B-16K and LLaMA-2-7B as the LLMs for evaluation, which are frozen during the optimization of QGC. To ensure stable and reproducible results, we employ greedy decoding and set the temperature to 0 in all experiments. 
Following~\citet{jiang2023longllmlingua}, we use LLaMA-2-7B-Chat as the external language model for Selective-Context and LongLLMLingua.
For QGC, both the query-guided context encoder and query-document reviewing layer consist of two Transformer encoder layers. 
All these layers and word embeddings are initialized with LLaMA-2-7B where MLP parameters are all fixed during training.
Our rationale behind this approach stems from our belief that the MLP plays a crucial role in knowledge retention, while our focus lies in adjusting the acquired knowledge based on query.
Thus, the trainable parameters in QGC are only 3.5\% of LongChat-13B-16K. Besides the ground-truth document, we concatenate 1-4 random documents to build the long context. We also randomly set the n-gram size from the candidate list (4, 6, 8, 10) for each training batch to make the compressor more robust. We train QGC on downstream datasets for 15 epochs, using a learning rate of 5e-5 with the Adam optimizer and batch size of 64.
During inference, we use the Cond.PPL proposed by~\citet{jiang2023longllmlingua} to sort retrieved documents for all compression-based methods and QGC, and set the $\epsilon$ as 0.35. 
Following~\cite{liu2023lost, bai2023longbench} the maximum generation tokens is 100 for NaturalQuestions, and 32 for both TriviaQA and HotpotQA.
All experiments are conducted on 8 NVIDIA A100 GPUs.

\begin{table}[t]
\fontsize{10}{12}\selectfont
\setlength{\tabcolsep}{1.1mm}
\centering
\begin{tabular}{lc}
\toprule
\textbf{Methods} & \textbf{Accuracy} \\
\midrule
QGC & \textbf{69.19} \\
\cdashline{1-2}
\quad \textit{w/o query-guided context encoder} & 50.36 \\
\quad \textit{w/o query-guided pooling layer} & 55.34 \\
\quad \textit{w/o query-document reviewing layer} & 64.14 \\
\quad \textit{w/o dynamically compressing strategy} & 62.15 \\
\bottomrule
\end{tabular}
\vspace{-2mm}
\caption{
The accuracy of ablation study on NaturalQuestions test set, where the target LLM is LongChat-13B.}
\vspace{-4mm}
\label{tab:ablation}
\end{table}

\paragraph{Main Results}
Table~\ref{tab:main-exp} reports the performance, compression ratios, and throughput of various methods or models on different datasets. Overall, QGC achieves higher compression ratios and greater throughput while achieving comparable or even better performance with LongLLMLingua. These results demonstrate that QGC can effectively compress context into shorter inputs.

Specifically, the performance and compression ratio of the reranker-based methods are limited because no compression operation is used within the document.
Compared to AutoCompressor and ICAE, our method achieves better accuracy with comparable compression ratios.
Compared with LongLLMLingua, QGC achieves average +5.03 and +12.87 performance improvements when using LongChat-13B and LLaMA-2-7B as the target LLMs. On average, the compression ratio and throughput of QGC are 2.75 times and 2.47 times that of LongLLMLingua on all datasets and target LLMs, respectively.

\paragraph{Ablation Study}

\begin{figure}[t!]
\centering
\subfigcapskip=-2.5mm
\subfigure[Compression Ratio for QGC]{
    \includegraphics[width=0.92\linewidth]{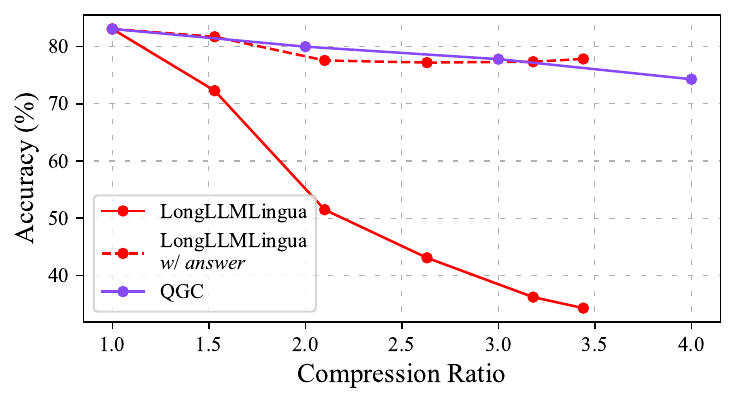}
    \label{fig:analysis-key-cr}
}

\vspace{-2.5mm}
\subfigure[Document Number for QGC]{
    \includegraphics[width=0.92\linewidth]{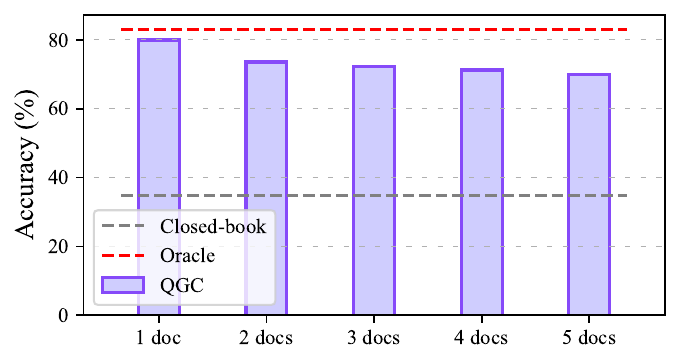}
    \label{fig:analysis-key-doc}
}
\vspace{-4mm}
\caption{The accuracy of QGC with varying compression ratios and number of documents, respectively.}
\vspace{-4mm}
\label{fig:analysis-key}
\end{figure}

To explore the effect of different components on QGC, we use LongChat-13B as the target LLM and introduce the following variants of QGC for ablation study: 1) \textit{w/o query-guided context encoder}. In this variant, the query and document are independently encoded; 2) \textit{w/o query-guided pooling layer}. When establishing this variant, we directly replace the weighted sum of token representations in each $n$-gram with their mean representation; 3) \textit{w/o query-document reviewing layer}. This variant no longer refines the compressed representations of $n$-grams; 4) \textit{w/o dynamically compressing strategy}.
We fix the $n$-gram size as 4 for comparable comparison.

As shown in Table~\ref{tab:ablation}, the absence of the query-document reviewing layer and dynamically compressing strategy lead to a 5.05 and 7.04 accuracy loss respectively. The more substantial loss is observed after removing the query-guided context encoder and query-guided pooling layer, resulting in a significant performance accuracy drop of 18.83 and 13.85 respectively, highlighting the importance of employing the query to guide compression.

\section{Analysis}
In this section, we conduct in-depth analyses to explore the performance of QGC in terms of key information loss, demonstration compression, detailed throughput and reranker impact. All analyses are conducted on NaturalQuestions with target LLM as LongChat-13B.

\paragraph{Key Information Loss in QGC}
As described in Section~\ref{sec:preliminary-study-key-info}, previous methods dramatically lose key information as the compression ratio increases. For comparison, we experiment with QGC using the same setting.

Compared to LongLLMLingua in Figure~\ref{fig:analysis-key-cr}, the performance of QGC only decreases 10\% as the compression ratio increases from 1x to 4x, and is even comparable to that of LongLLMLingua containing the correct answer in the compressed result. As seen in Figure~\ref{fig:analysis-key-doc}, we observe that the performance of QGC slightly degrades with more documents, which is only a 12\% decrease with 4 documents (27\% for AutoCompressor). These results demonstrate that QGC can effectively retain key information even in much longer context and higher compression ratio scenarios.

\paragraph{Demonstration Compression for In-Context Learning}

\begin{table}[t]
\fontsize{10}{12}\selectfont
\setlength{\tabcolsep}{1.1mm}
\centering
\begin{tabular}{lcccc}
\toprule
\multirow{2}{*}{\textbf{Methods}} & \multicolumn{2}{c}{\textbf{SST-2}} & \multicolumn{2}{c}{\textbf{GSM8K}} \\
\cmidrule(r){2-3} \cmidrule(r){4-5}
& Acc & CR & Acc & CR \\
\midrule
Original Prompt & 92.4 & 1.0x & \textbf{14.48} & 1.0x \\
LongLLMLingua & - & - & 5.91 & 3.9x \\
AutoCompressor & 94.2 & 15.0x & 6.68 & \textbf{13.6x} \\
\midrule
QGC & \textbf{94.8} & \textbf{23.3x} & 14.18 & 13.4x \\
\bottomrule
\end{tabular}
\vspace{-2mm}
\caption{Experimental results on SST-2 and GSM8K datasets, where the target LLM is LLaMA-2-7B.}
\vspace{-4mm}
\label{tab:icl}
\end{table}

To further validate the effectiveness of QGC in a broader context, we conduct experiments on both SST-2 and GSM8K datasets. We adopt the approach of previous studies~\cite{chevalier2023adapting, wei2022chain} which utilizing demonstrations as the document, while maintaining consistency with their experimental setup.
The results in Table~\ref{tab:icl} reveals notable insights. On the SST-2 dataset, our method surpasses autocompressor in both compression ratio and accuracy. Meanwhile, on the GSM8K dataset, our accuracy performance remains on par with the original prompt at the same compression ratio as autocompressor. This suggests that QGC strikes an excellent balance between model performance and compression ratio. These results showcases QGC's proficiency in preserving information from demonstrations and fostering the in-context learning capacity of the target LLM.

\paragraph{Detailed Throughput Evaluation}

\begin{figure}[t!]
\centering
\includegraphics[width=0.96\linewidth, trim=0 3mm 0 3mm]{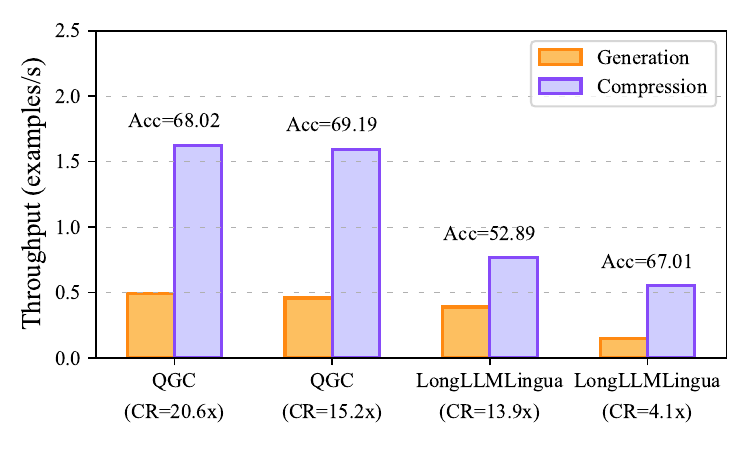}
\vspace{-3mm}
\caption{The accuracy, compression throughput, and generation throughput of QGC and LongLLMLingua.}
\vspace{-5mm}
\label{fig:throughput}
\end{figure}

To evaluate the throughput of various methods or models, encompassing both compression and generation, we perform testing on a single A100-80G GPU. 

The results presented in Figure~\ref{fig:throughput} indicate that QGC is obviously higher than LongLLMLingua in both compression throughput and generation throughput. Moreover, by adjusting the hyper-parameter $\epsilon$ (See Equation~\ref{eq:compression-rate}) to increase the compression ratio, QGC can achieve a higher compression ratio while minimizing the impact on LLM performance and further improving throughput. Furthermore, our higher compression ratios lead to shorter LLM input, which also significantly improves the generation throughput of the target LLM. As for LongLLMLingua, since it additionally introduces LLaMA-2-7B for compression, the compression throughput is significantly lower than ours. Besides, although LongLLMLingua can also improve compression ratio by adjusting hyper-parameters, its performance will significantly drop, while QGC still maintains excellent performance.

\paragraph{Impact of Different Rerankers}
The compression ratio for each document is determined by the corresponding correlation with the query obtained by a reranker. 
Here, we analyze the impact of using different rerankers in this process. In addition to the three methods introduced in reranker-based methods, we also include BM25 and Gzip~\cite{jiang-etal-2023-low} for comparison.

\begin{figure}[t!]
\centering
\includegraphics[width=0.98\linewidth]{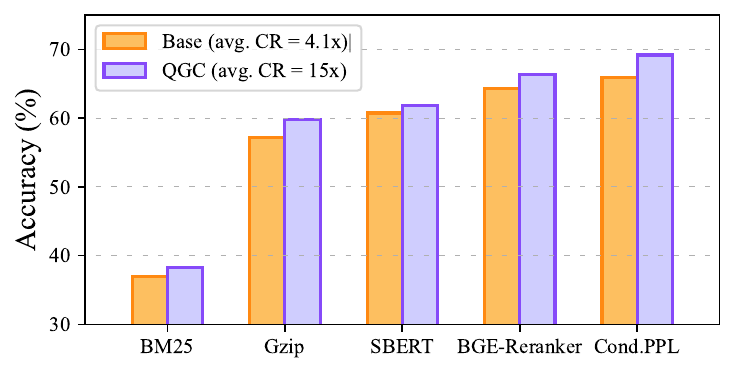}
\vspace{-3mm}
\caption{The performance of QGC using different rerankers. ``Base'' represents the performance of each reranker to be used for compression. The performance (Recall) of rerankers: Cond.PPL > BGE-Rererank > SBERT (Sentence-BERT) > Gzip > BM25.}
\vspace{-5mm}
\label{fig:analysis-reranker}
\end{figure}

Experimental results are shown in Figure~\ref{fig:analysis-reranker}.
It can be found that QGC performs better with more competitive rerankers.
Besides, compared with directly using rerankers for compression, QGC not only achieves an average 2.65 times higher compression ratio but also maintains lossless or even improved performance.

\section{Related Work}
\paragraph{Long Context for LLMs}
Recently, there have been a lot of studies focusing on expanding the context length of LLMs \cite{press2021train,peng2023yarn,bertsch2023unlimiformer}. Existing efforts primarily involve gradually increasing the window size during pre-training \cite{nijkamp2023xgen}, interpolating position embeddings \cite{chen2023extending}, and modifying the attention mechanism \cite{ding2023longnet}.
Unlike these works, we do not directly aim to expand the context window of LLMs. Hence, the QGC that we proposed can complement these techniques by enabling LLMs to access a broader context with reduced cost and shorter latency.

\paragraph{Retrieval-augmented LMs}
Combined with a standalone retriever to augment LMs are gaining popularity for benefiting various knowledge-intensive tasks. Previous studies have achieved remarkable results in improving perplexity~\cite{wang2023shall}, factual accuracy~\cite{nakano2022webgpt}, downstream task performance~\cite{izacard2022atlas}, and in-context learning~\cite{huang2023raven}. Besides, many works focus on cooperating LLMs and retrieved documents, such as reranking retrieved documents~\cite{mao-etal-2021-reader} and discarding irrelevant documents~\cite{mallen-etal-2023-trust}.
QGC is also a retrieval augmentation method for LLMs, which efficiently compresses the retrieved documents into shorter inputs while maintaining no significant performance degradation.

\paragraph{Context Compression}
With the growing context length in LLMs, the demand for higher efficiency, lower cost, and reduced latency has attracted much attention. As a promising solution, compression techniques can be broadly categorized into two types: black-box compression~\cite{xu2023recomp} and white-box compression~\cite{wang2023learning}. Black-box compression primarily involves token pruning based on different importance measures, such as self-information~\cite{li2023compressing} and the LLM perplexity~\cite{jiang-etal-2023-llmlingua, jiang2023longllmlingua}. 
On the other hand, white-box compression focuses on generating summarization or compressing the context into soft prompt through fine-tuning or Low-Rank Adaptation (LoRA).
For instance, \citet{wang2023learning} autoregressively generates filtered content and fine-tunes target LLM to use it for generation. \citet{mu2023learning} trains LLMs to compress instructions into concise key-value attention prefixes. \citet{chevalier2023adapting} recursively compresses lengthy text into summary vectors, while \citet{ge2023context} generates memory slots to represent the original context.
Compared with the above-mentioned compression studies, QGC's design fully takes into account the query, which leads to the enhanced retention of key information, higher compression ratios, higher throughput, and improved overall performance.

\section{Conclusion and Future Work}
In this paper, we have presented a query-guided compressor QGC for LLMs to solve the loss of key information under high compression ratios. It consists of four essential components: query-guided context encoder, query-guided pooling layer, query-document reviewing layer, and semantic alignment layer. In addition, we also propose a dynamically compressing strategy during inference.
Extensive experiments on multi-document QA tasks demonstrate that QGC outperforms previous state-of-the-art compression methods in both accuracy and compression ratios. Analyses reveal that this is primarily due to our retention of key information throughout the compression process.

In the future, we aim to validate our approach on more advanced LLMs, while also expanding its application to additional tasks like document summarization. Besides, we will try to further improve our approach by combining previous studies~\cite{zhang2018simplifying,hu2022lora,zhang2022aan+,zhang-etal-2022-moefication}.

\section*{Limitations}
QGC is a white-box compressor that necessitates access to the internal parameters of LLMs, which restricts its applicability. 
Furthermore, we have solely validated the effectiveness of QGC on QA and ICL task, and its performance on other tasks that differ significantly from QA task, such as summarization, remains to be verified.

\section*{Acknowledgements}
The project was supported by National Key R\&D Program of China (No. 2022ZD0160501), National Natural Science Foundation of China (No. 62276219), and the Public Technology Service Platform Project of Xiamen (No. 3502Z20231043). We also thank the reviewers for their insightful comments.

\bibliography{anthology,custom}

\appendix
\newpage

\section{Instructions Used in QGC}
The following are the instructions we used after referring to the existing studies~\cite{liu2023lost} and testing.
\begin{itemize}
    \item \textbf{NaturalQuestions}: \textit{Write a high-quality answer for the given question using only the provided search results(some of which might be irrelevant)}.
    \item \textbf{TriviaQA} \& \textbf{HotpotQA}: \textit{Using only the provided search results (some of which might be irrelevant), answer the following question with one or few words}.
\end{itemize}

\end{document}